\documentclass[letterpaper]{article} 
\usepackage{aaai2027}[] 
\usepackage[hyphens]{url} 
\usepackage{graphicx} 
\usepackage{natbib} 
\usepackage{caption} 
\usepackage{amsmath}
\usepackage{amssymb}
\usepackage{booktabs}
\usepackage{multirow}

\nocopyright
\newcommand{\best}[1]{\textbf{#1}}
\newcommand{\improved}[1]{\textbf{#1}\,$\uparrow$}

\title{Improving Auto-Design of Neural PDE Solvers with a Domain-Specific Language}
\author{
Shengxin Kong\textsuperscript{\rm 1},
Liwen Xu\textsuperscript{\rm 1}\corresponding,
and Jingwen Fu\textsuperscript{\rm 2}\corresponding
}
\affiliations{
\textsuperscript{\rm 1}School of Sciences, North China University of Technology, Beijing, China\\
\textsuperscript{\rm 2}Beijing Zhongguancun Academy, Beijing, China\\
kongshengxin709@gmail.com,
xulw@ncut.edu.cn,
fujingwen@bza.edu.cn
}

\begin{document}

\maketitle

\begin{abstract}
Neural PDE solver auto-design is fundamentally a search-space representation problem. In the space of unrestricted Python programs, valid solvers form an extremely sparse subset: most candidate programs are syntactically incorrect, semantically incompatible, or numerically unstable. Direct code generation therefore forces an LLM to spend most of its search capacity navigating implementation failures rather than reasoning about solver quality.
ADSL-PDE\footnotemark{} addresses this challenge by introducing a structured search state between solver concepts and executable code. It represents the functional decisions that determine a neural PDE solver—architecture, physical constraints, objectives, sampling, and optimization—while abstracting away low-level implementation details. A deterministic compiler maps each valid search state to an executable solver. In effect, ADSL-PDE reshapes the search space: it removes large regions of invalid programs, increases the density of meaningful candidates, and preserves the compositional freedom needed to discover previously unseen designs. Solver evolution can thus operate over design decisions rather than code artifacts.
Built on this representation, our evolutionary agent iteratively proposes, evaluates, and refines solver search states using empirical feedback. Across multiple PDE benchmarks, ADSL-PDE improves both search efficiency and optimization stability, achieving more than 52\% performance improvement within the first ten evolution iterations. These results suggest a broader principle for LLM-driven auto-design: effective agents do not merely require stronger reasoning, but a search representation that concentrates exploration on valid and consequential decisions.
\end{abstract}

\footnotetext{The source code is available at
\url{https://github.com/Super-KongCC/Improving-Auto-Design-of-Neural-PDE-Solvers-with-a-Domain-Specific-Language}.}


\section{Introduction}

Neural partial differential equation (PDE) solver auto-design has emerged as an important research direction in AI for Science, aiming to automatically discover effective solver designs while reducing reliance on human expertise \cite{raissi2019physics,lu2021deeponet,li2021fourier}. This task involves searching over a structured design space that includes network architectures, loss functions, sampling strategies, and training hyperparameters. Existing approaches can be broadly divided into search-based and agent-based methods. Search-based methods use neural architecture search, evolutionary algorithms, or Bayesian optimization to identify solver configurations within predefined search spaces \cite{elsken2019nas,zoph2017nas,real2019evolution}. Although effective in some cases, these methods are limited by manually designed search spaces and may not capture the full range of neural PDE solver designs. Agent-based methods instead use the reasoning and code-generation abilities of large language models to iteratively generate and improve solvers \cite{yao2023react,schick2023toolformer,wuwu2025pinnsagent}. However, most of them directly generate Python code, requiring agents to handle both solver design and implementation details, such as framework syntax, function calls, and dependency management. This greatly increases the search complexity and can lead to invalid programs and unstable training. Therefore, search-based methods are limited by overly restricted spaces, while agent-based methods face overly large and implementation-heavy spaces. Neither provides an effective representation for reliable and flexible neural PDE solver auto-design.

\begin{figure*}[t]
\centering
\includegraphics[width=0.90\linewidth]{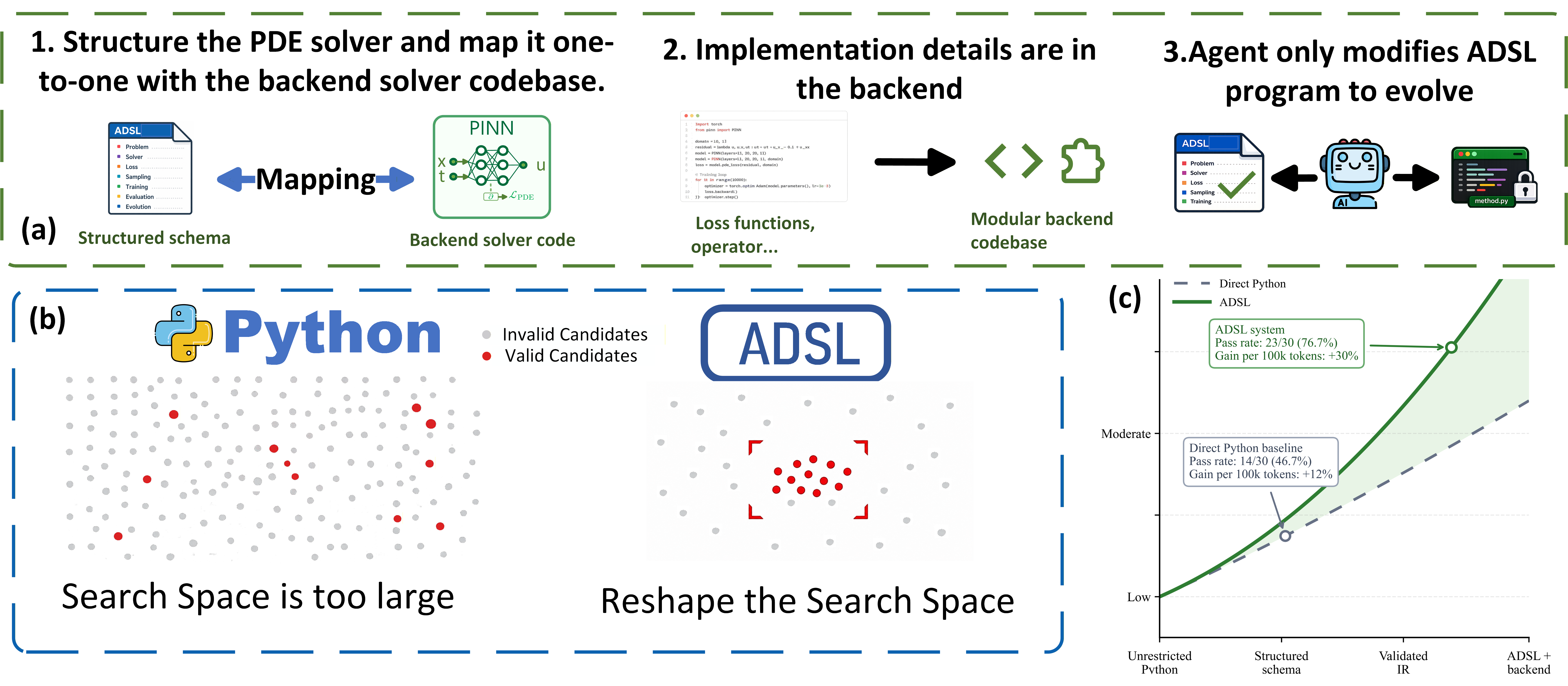}
\caption{Reshaping the neural PDE solver search space with ADSL-PDE.
(a) ADSL-PDE represents solver designs with a structured schema, maps them to a modular backend, and restricts the agent to editing ADSL programs.
(b) This language design removes unproductive regions of the unrestricted Python space, yielding a smaller and denser search space of valid solver candidates.
(c) The resulting structured search space enables more effective optimization; quantitative results are reported in Table~\ref{tab:key_design_ablation}.}
\label{fig:framework_overview}
\end{figure*}

Domain-specific languages (DSLs) have long been used in software engineering to provide compact and structured representations for specialized tasks \cite{fowler2010dsl,parr1998dsl}. Inspired by this idea, we propose ADSL-PDE (Agent-oriented Domain-Specific Language for PDE Solver Evolution), a DSL tailored for neural PDE solver design, to address the representation limitations of existing agent-based auto-design approaches. As illustrated in Figure~\ref{fig:framework_overview}, the core idea of ADSL-PDE is to reshape the search space for neural PDE solver auto-design. Unlike approaches that directly generate Python code, ADSL-PDE decouples solver design semantics from implementation logic. The language frontend explicitly represents high-level design components, including network architectures, physical constraints, loss formulations, and training strategies, while a deterministic backend translates these specifications into executable solvers. This design allows agents to evolve neural PDE solvers by modifying structured design scripts rather than repeatedly manipulating complex and error-prone implementation code.

Our contributions are summarized as follows:

(1) We identify representation as a key bottleneck in neural PDE solver auto-design and highlight the critical role of language design in enabling efficient agent-based solver evolution.

(2) We propose ADSL-PDE, an agent-oriented domain-specific language and evolutionary framework that transforms neural PDE solver design from unrestricted code generation into structured design-space exploration.

(3) We validate ADSL-PDE on multiple PDE benchmarks, where it consistently improves search efficiency and achieves over 52\% performance gains within the first ten evolution iterations.

\begin{figure}[t]
    \centering
    \includegraphics[width=0.85\columnwidth]{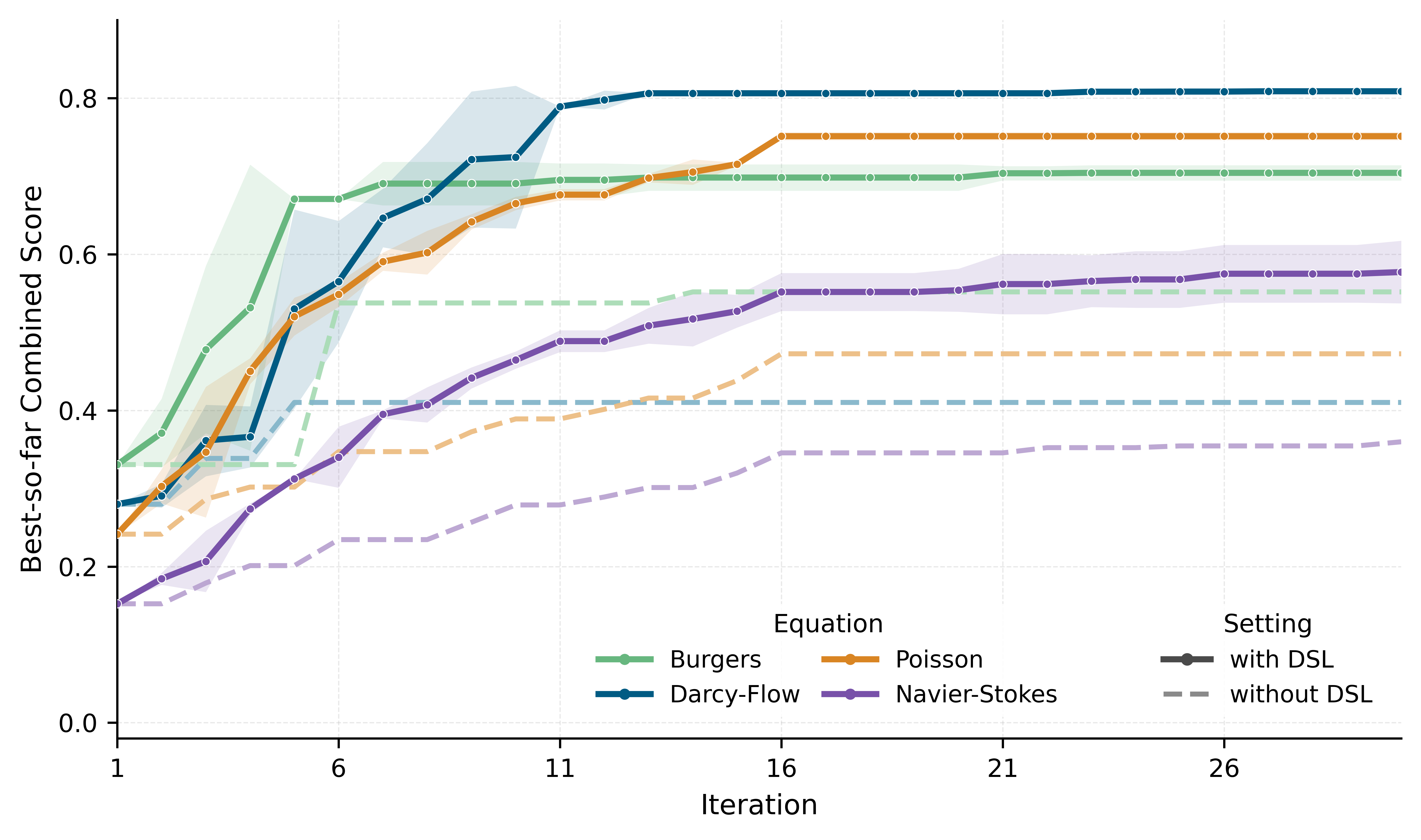}
 \caption{Combined scores on Darcy-Flow and Burgers. Solid curves with shaded bands denote DSL-based results, while lighter dashed curves denote results without DSL.}
    \label{fig:combined_scores}
\end{figure}

\section{Related Work}

\subsection{Neural PDE Solvers}

Neural PDE solvers are widely used in scientific machine learning. Physics-informed neural networks incorporate governing equations into training objectives \cite{raissi2019physics}, while neural operators such as DeepONet, FNO, and U-NO learn mappings between PDE inputs and solutions \cite{lu2021deeponet,li2021fourier,rahman2022uno}. Graph-based simulators support irregular meshes and complex geometries \cite{pfaff2021meshgraphnets}, and PDEBench and PDENNEval provide systematic evaluations of these solver families \cite{takamoto2022pdebench,wei2024pdenneval}. However, their performance is strongly task-dependent: PINNs are sensitive to gradient imbalance, loss weighting, sampling, and optimization \cite{wang2021gradientpathologies,wang2023expertpinn}, whereas neural operators and graph-based methods rely on different data representations and inductive biases. We therefore focus on representing and exploring solver designs across multiple solver families rather than proposing another fixed architecture.

\subsection{Automated Design of Neural PDE Solvers}

Neural solver design has been automated through architecture search, evolutionary algorithms, and hyperparameter optimization \cite{elsken2019nas,zoph2017nas,real2019evolution}. PDE-specific methods such as Auto-PINN, NAS-PINN, and PINN-DARTS search predefined PINN architecture or training spaces \cite{wang2022autopinn,wang2024naspinn,li2025pinndarts}. More recent methods, including PINNsAgent, Lang-PINN, and CodePDE, use LLMs to generate or refine PDE solver implementations \cite{wuwu2025pinnsagent,he2025langpinn,li2025codepde}. AlphaEvolve and OpenEvolve further combine LLM-based mutation with automated evaluation and population selection \cite{novikov2025alphaevolve,sharma2025openevolve}. Traditional search methods are constrained by predefined spaces, whereas code-generating agents must handle both solver design and low-level implementation. ADSL-PDE instead evolves structured solver specifications that can be validated and compiled deterministically.

\subsection{Agent-Oriented Languages and Structured Representations}

Domain-specific languages provide compact abstractions for specialized tasks \cite{fowler2010dsl}, and recent studies suggest that representation can influence LLM reasoning and internal feature activation \cite{yang2026shapingschema,zhang2025srllm}. LMQL introduces constraints and control flow for language-model generation \cite{beurerkellner2023lmql}, while PAL and Logic-LM translate natural-language problems into executable programmatic or symbolic forms \cite{gao2023pal,pan2023logiclm}. Structured agent interfaces include CodeStruct for syntax-aware code manipulation \cite{kim2026codestruct}, ACDL for agent-context construction \cite{pelegpelc2026acdl}, and Extended Scenic for executable driving scenarios \cite{safa2026llm}. Unlike these systems, ADSL-PDE integrates domain semantics, static validation, backend compilation, evaluator feedback, and iterative optimization, using a PDE-specific language as both the representation and action space for neural solver evolution.

\section{Method}

\subsection{PDE Problem Formulation}

We consider a general PDE task defined on a spatial-temporal domain
$\Omega \times [0,T]$, with target solution
$u:\Omega \times [0,T]\rightarrow \mathbb{R}^{d_u}$. The problem is written as
\begin{equation}
\begin{aligned}
\mathcal{N}[u](x,t) &= f(x,t),
&& (x,t)\in\Omega\times[0,T],\\
\mathcal{B}[u](x,t) &= g(x,t),
&& (x,t)\in\partial\Omega\times[0,T],\\
u(x,0) &= u_0(x),
&& x\in\Omega,
\end{aligned}
\end{equation}
where $\mathcal{N}$ and $\mathcal{B}$ denote the differential and boundary
operators, respectively.

A neural PDE solver is determined by a coupled set of design choices rather
than by its network architecture alone. We represent a solver design as the
structured schema
\begin{equation}
z =
(\mathcal{T},\mathcal{F},\mathcal{A},\mathcal{C},
\mathcal{S},\mathcal{R},\mathcal{E}),
\end{equation}
where $\mathcal{T}$ denotes the PDE task, $\mathcal{F}$ the solver family,
$\mathcal{A}$ the neural architecture, $\mathcal{C}$ the physical constraints,
$\mathcal{S}$ the sampling strategy, $\mathcal{R}$ the training recipe, and
$\mathcal{E}$ the evaluation protocol. Given a task $\mathcal{T}$, solver
auto-design searches for a valid schema
$z\in\mathcal{Z}^{\mathrm{valid}}_{\mathcal{T}}$ with strong validation
performance. ADSL-PDE represents and evolves solver designs at this schema
level rather than directly searching over implementation code.

\subsection{Overview of ADSL-PDE}

The framework of ADSL-PDE consists of two coupled components: a language system and an evolution system. The language system defines how solver designs are expressed, checked, and executed, while the evolution system defines how an LLM agent modifies solver designs under feedback.

Given an initial DSL program, the LLM agent proposes a modified program by editing solver-level fields rather than low-level implementation code. The parser converts the program into a typed intermediate representation (IR), where solver components and their dependencies are explicitly represented. The verifier then checks whether the IR satisfies task-level, solver-level, and backend-level constraints. Only verified candidates are passed to the compiler, which instantiates an executable neural PDE solver under a deterministic backend.

The compiled solver is first evaluated through a quick training stage. Promising candidates are selected for full training, while invalid or poorly performing candidates are converted into structured feedback. This feedback is aligned with the DSL fields edited by the agent, allowing subsequent modifications to target meaningful solver-design decisions such as loss weighting, sampling density, architecture depth, activation functions, or optimization schedules. In this way, ADSL-PDE transforms agent-based solver evolution from unrestricted code generation into verifiable solver-schema search.

\begin{table*}[t]
\centering
\small
\setlength{\tabcolsep}{1.5mm}
\begin{tabular}{lcccccccc}
\toprule
& \multicolumn{3}{c}{Function-learning methods}
& \multicolumn{4}{c}{Operator-learning methods}
& \multicolumn{1}{c}{Ours} \\
\cmidrule(lr){2-4}
\cmidrule(lr){5-8}
\cmidrule(lr){9-9}
PDE
& PINN
& DFVM
& RFM
& PINO
& U-NO
& FNO
& MPNN
& ADSL-PDE \\
\midrule
1D Advection
& 0.9971
& 0.1042
& 0.9767
& 0.0191
& \textbf{0.0069}
& 0.0128
& 0.0451
& 0.0094 \\

1D Diffusion-Reaction
& 0.0470
& 0.2434
& 0.5769
& 0.0163
& 0.0036
& 0.0098
& 0.0039
& \textbf{0.0012} \\

1D Burgers
& 0.1736
& 0.0347
& 1.1146
& 0.1206
& 0.0601
& 0.0603
& 0.1143
& \textbf{0.0022} \\

1D Diffusion-Sorption
& 0.1920
& 0.5967
& 0.4296
& 0.0078
& 0.0012
& 0.0024
& 0.0029
& \textbf{0.0008} \\

1D Allen-Cahn
& 0.9999
& 1.4700
& 247.86
& 0.0062
& 0.0019
& 0.0041
& 0.0033
& \textbf{0.0003} \\

1D Cahn-Hilliard
& 0.1520
& 1.1887
& 10.579
& 0.0259
& 0.0041
& 0.0053
& 0.0554
& \textbf{0.0033} \\

\midrule

2D Burgers ($u$)
& 1.5809
& 0.0117
& 39.859
& 0.0298
& 0.0089
& 0.2201
& 0.0115
& \textbf{0.0016} \\

2D Darcy Flow
& --
& --
& --
& 0.0702
& 0.0742
& 0.1328
& --
& \textbf{0.0042} \\

2D Shallow-Water
& 0.0120
& 0.0114
& 0.0601
& 0.0166
& 0.0025
& 0.0036
& 0.0011
& \textbf{0.0005} \\

2D Allen-Cahn
& 0.4125
& 0.1363
& 588.58
& 0.0191
& 0.0054
& 0.0084
& 0.0075
& \textbf{0.0047} \\

2D Black-Scholes
& 0.2365
& 0.0357
& 0.5423
& 0.0447
& 0.0021
& 0.0072
& 0.0015
& \textbf{0.0012} \\

\midrule

Geo. Mean
& 2.82E-01
& 9.77E-02
& 4.02E+00
& 2.45E-02
& 6.06E-03
& 1.71E-02
& 8.54E-03
& \textbf{1.16E-03} \\

\bottomrule
\end{tabular}
\caption{Comparison with manually designed function-learning and
operator-learning baselines. The reported metric is MSE or relative
$L_2$ error, depending on the benchmark, and lower values indicate
better performance. The geometric mean is computed over available
entries, excluding missing results. The best result for each benchmark
is shown in bold.}
\label{tab:manual_baselines}
\end{table*}

\subsection{ADSL Language Design}

ADSL-PDE is designed as an agent-facing language for neural PDE solver design. Its purpose is not merely to provide a configuration format, but to reshape the action space of the LLM agent. Instead of allowing the agent to generate arbitrary Python code, ADSL exposes a compact set of solver-level primitives that correspond to the components of the solver schema defined above. A complete ADSL program contains typed blocks for the PDE task, solver family, network architecture, physical constraints, sampler, trainer, evaluator, and evolution policy.

The front-end language follows a principle of controlled expressiveness. It is more restrictive than Python because the agent cannot define arbitrary functions, import external libraries, modify backend training loops, or bypass the compiler. At the same time, it is more expressive than a fixed hyperparameter table because it allows the agent to modify meaningful PDE-solver decisions, including architecture depth and width, activation functions, residual and boundary loss weights, adaptive sampling strategies, optimizer choices, learning-rate schedules, and training stages. Moreover, at a frequency controlled by a dedicated hyperparameter, the agent can synthesize new solver implementations by combining design elements from high-performing solvers and, after validation, register them in the backend code library. This design removes implementation-level degrees of freedom while preserving solver-level flexibility.

The language primitives are aligned with PDE solver semantics. The \texttt{task} block specifies the equation, domain, variables, boundary and initial conditions, and evaluation metric. The \texttt{solver} block determines the solver family and automatic differentiation mode. The \texttt{network} block defines the neural parameterization. The \texttt{physics} block specifies residual, boundary, data, and constraint losses. The \texttt{sampler} block controls how interior, boundary, and data points are selected. The \texttt{trainer} block describes the optimization recipe, while the \texttt{evaluator} and \texttt{evolution} blocks define model selection and editable mutation scopes.

After parsing, an ADSL program is converted into an IR process. The IR serves as the canonical representation of the solver schema. It removes superficial syntactic variations in the front-end program and represents the solver as a dependency graph of typed components. This representation makes the dependencies among solver components explicit and enables static checking before execution.

The verifier enforces the validity of the solver schema before compilation. It checks whether the PDE task is complete, whether component input-output types are consistent, whether tensor shapes and variable references are compatible, whether the selected solver family can support the requested losses and derivative orders, and whether every IR primitive has a corresponding backend implementation. For instance, if a candidate refers to an undefined physical variable, uses an unsupported sampler for a given domain, or selects a solver configuration that cannot compute the required derivatives, the verifier rejects the candidate before training. The rejection is not returned as a generic runtime failure, but as diagnostic feedback tied to the corresponding DSL field.

The backend compiler grounds a verified IR into an executable solver. The compiler instantiates neural modules, automatic differentiation residuals, loss functions, samplers, optimizers, schedulers, training loops, checkpointing, and evaluation routines according to the verified IR. It is deterministic: the same IR always produces the same solver implementation. Importantly, the compiler does not perform hidden search, tune hyperparameters, or introduce task-specific heuristics beyond the DSL specification. This separation is essential for attributing the improvement to language-mediated solver search rather than uncontrolled backend optimization.

Through this language design, ADSL-PDE provides a middle ground between fixed search spaces and unrestricted code generation. It gives the agent access to expressive PDE solver-design primitives while enforcing type, feasibility, and executability constraints before training. This language system eliminates unproductive regions of the agent's search space, thereby improving the efficiency of agent-driven code evolution.

\subsection{Language-Guided Evolution Framework}

We build a language-guided evolutionary framework on top of ADSL-PDE.
The population is divided into method islands, where each island
corresponds to a neural solver family, such as PINN, FNO, U-NO, and so on. Each solver family instantiates the unified solver schema with
its own architecture, loss construction, sampling mechanism, data
interface, and training procedure. The island therefore defines a
family-specific search space while maintaining a common representation
across heterogeneous solvers.

In each evolution round, the framework selects an island and a parent
ADSL program. Conditioned on the parent, the island constraints, and
historical feedback, the LLM agent modifies valid solver-level fields
within that family. The resulting program is parsed into a typed IR,
verified against both the unified schema and the selected solver family,
and then compiled into an executable solver. Invalid candidates are
rejected before training and returned as structured diagnostic feedback.

Valid candidates are first evaluated with a limited quick-training
budget. Promising candidates are then selected for full training, and
the island populations are updated according to validation performance.
The framework returns numerical feedback, including validation error and
training stability, together with structural and diagnostic feedback
linked to the modified ADSL fields. This allows the agent to revise
solver designs rather than repair low-level code.

\begin{figure}
    \centering
    \includegraphics[width=1.05\linewidth]{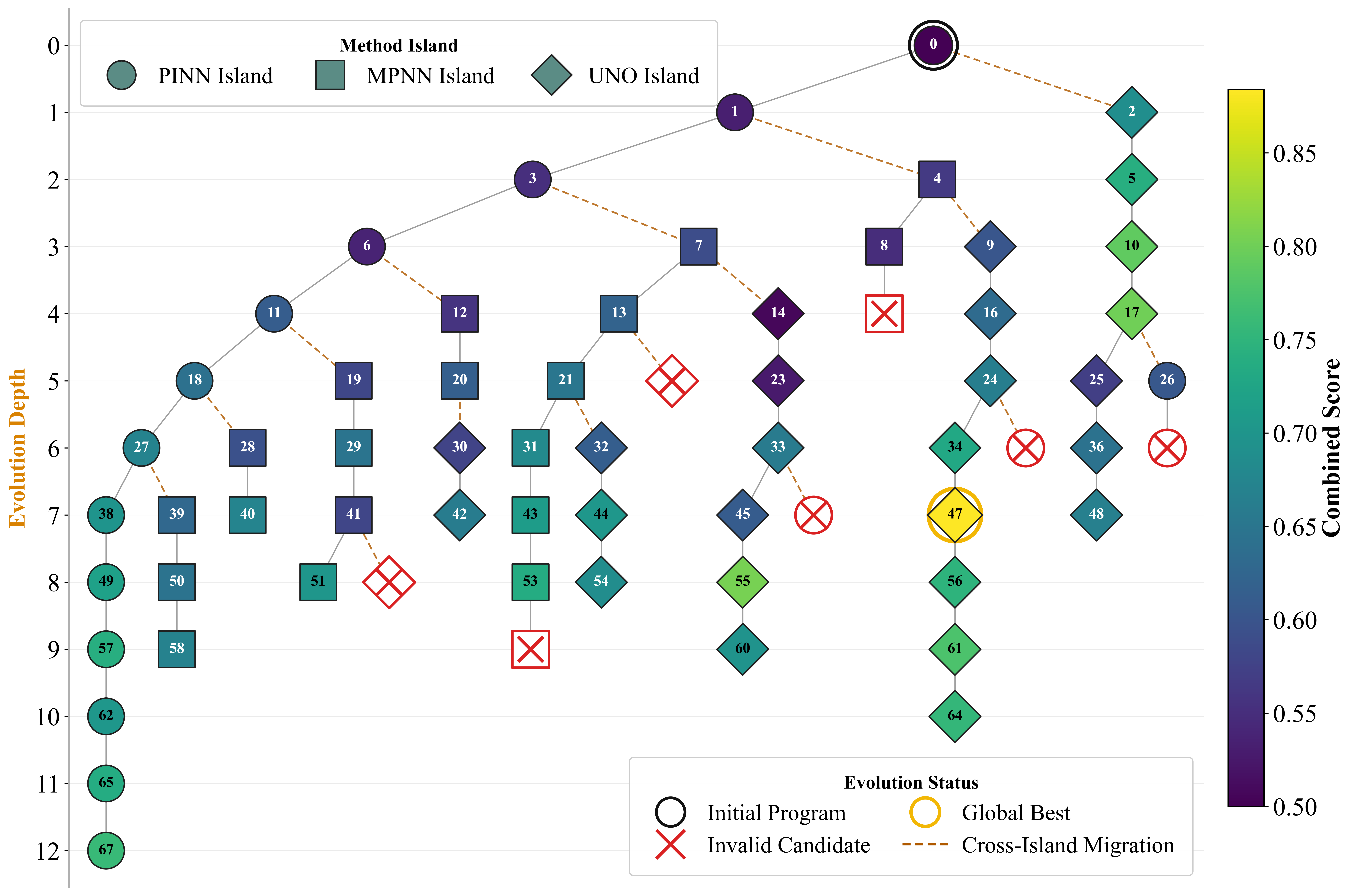}
    \caption{The complete evolution path of the Burgers equation under ADSL-PDE.}
    \label{fig:evolution}
\end{figure}

The framework also supports controlled cross-island innovation. At a
frequency specified by an evolution hyperparameter, the agent can combine
compatible components from high-performing solvers to create a new solver
implementation. After validation, the implementation is registered in
the backend library and can participate in subsequent evolution as a new
method island. The code selection and evolution logic can be swapped with different evolution frameworks, such as OpenEvolve \cite{openevolve}. Figure~\ref{fig:evolution} illustrates the evolutionary genealogy of candidate neural solvers for the Burgers equation across the PINN, MPNN, and UNO method islands. The framework explores diverse solver designs through within-island evolution and cross-island migration, while filtering invalid candidates and retaining the globally best program.

\section{Experiments}

We evaluate ADSL-PDE from three perspectives: final solver accuracy, comparison with search-based and LLM-agent baselines, and representation-level reliability. The main experiments are designed to test whether a structured agent-facing language can improve neural PDE solver auto-design beyond manually designed solvers, direct search baselines, and prompt-driven agent methods.

\subsection{Experimental Setup}

\paragraph{Benchmarks.}
We evaluate ADSL-PDE on a diverse set of PDE benchmarks covering one-dimensional, two-dimensional, three-dimensional, and high-dimensional tasks. The benchmark set includes advection, diffusion-reaction, Burgers, diffusion-sorption, Allen--Cahn, Cahn--Hilliard, Darcy flow, shallow-water, Black-Scholes-Barenblatt, heat, Navier--Stokes, Poisson, and Kuramoto--Sivashinsky equations. These tasks cover convection-dominated dynamics, nonlinear diffusion, multi-field coupled systems, elliptic equations, and high-dimensional PDEs.

\paragraph{Baselines.}
We compare ADSL-PDE with two groups of baselines. The first group consists of manually designed neural PDE solvers, including function-learning methods and operator-learning methods. Specifically, we compare with PINN \cite{raissi2019physics}, DFVM \cite{cen2024dfvm}, RFM \cite{chen2022rfm}, PINO \cite{li2024pino}, U-NO \cite{rahman2022uno}, FNO \cite{li2021fourier} and MPNN \cite{pfaff2021meshgraphnets}. This comparison evaluates whether ADSL-PDE can discover solver designs that are competitive with human-designed neural PDE solvers. The second group consists of search-based and LLM-agent-based auto-design baselines, including RandomAgent, BayesianAgent, PINNsAgent \cite{wuwu2025pinnsagent}, Lang-PINN \cite{he2025langpinn}, and PINNacle \cite{hao2024pinnacle} . This comparison evaluates whether the proposed language representation improves agent-based solver search.

\paragraph{Metrics and implementation.}
We report mean squared error (MSE) or relative $L_2$ error according to the metric used by each benchmark. Lower values indicate better predictive accuracy. To summarize performance across heterogeneous tasks with different error scales, we additionally report the geometric mean over available entries. For representation-level reliability, we report valid candidate rate (VCR), search efficiency (SE), and token-normalized search efficiency (Token-SE). Unless otherwise specified, the evolutionary search uses DeepSeek-V4-Pro as the LLM backbone. All representation ablations use the same PDE tasks, search budget, and backend training protocol. For ADSL-PDE, the LLM agent only modifies the front-end DSL program; each candidate is parsed into IR, statically verified, compiled by the deterministic backend, and evaluated by the same training and validation pipeline.

\subsection{Main Results against Manually Designed Solvers}

Table~\ref{tab:manual_baselines} compares ADSL-PDE with manually designed neural PDE solvers. ADSL-PDE achieves the best result on all listed benchmark entries and obtains the lowest geometric mean error. The result indicates that the proposed language representation does not sacrifice solver expressiveness: although the agent is restricted to editing structured DSL programs rather than arbitrary Python code, it can still discover high-quality solver designs across different PDE families.

The improvement is especially clear on nonlinear and coupled PDEs. For example, on 1D Burgers, ADSL-PDE reduces the error from $6.01\times10^{-2}$ for U-NO and $6.03\times10^{-2}$ for FNO to $2.20\times10^{-3}$. On 2D Burgers, ADSL-PDE achieves $1.60\times10^{-3}$ for the $u$ component and $9.00\times10^{-4}$ for the $v$ component. On 2D Darcy Flow, where operator-learning methods are usually strong baselines, ADSL-PDE obtains $4.20\times10^{-3}$, substantially lower than FNO, U-NO, and U-Net. These results suggest that ADSL-PDE provides an effective design interface for solver evolution.

\begin{figure}
    \centering
    \includegraphics[width=0.95\linewidth]{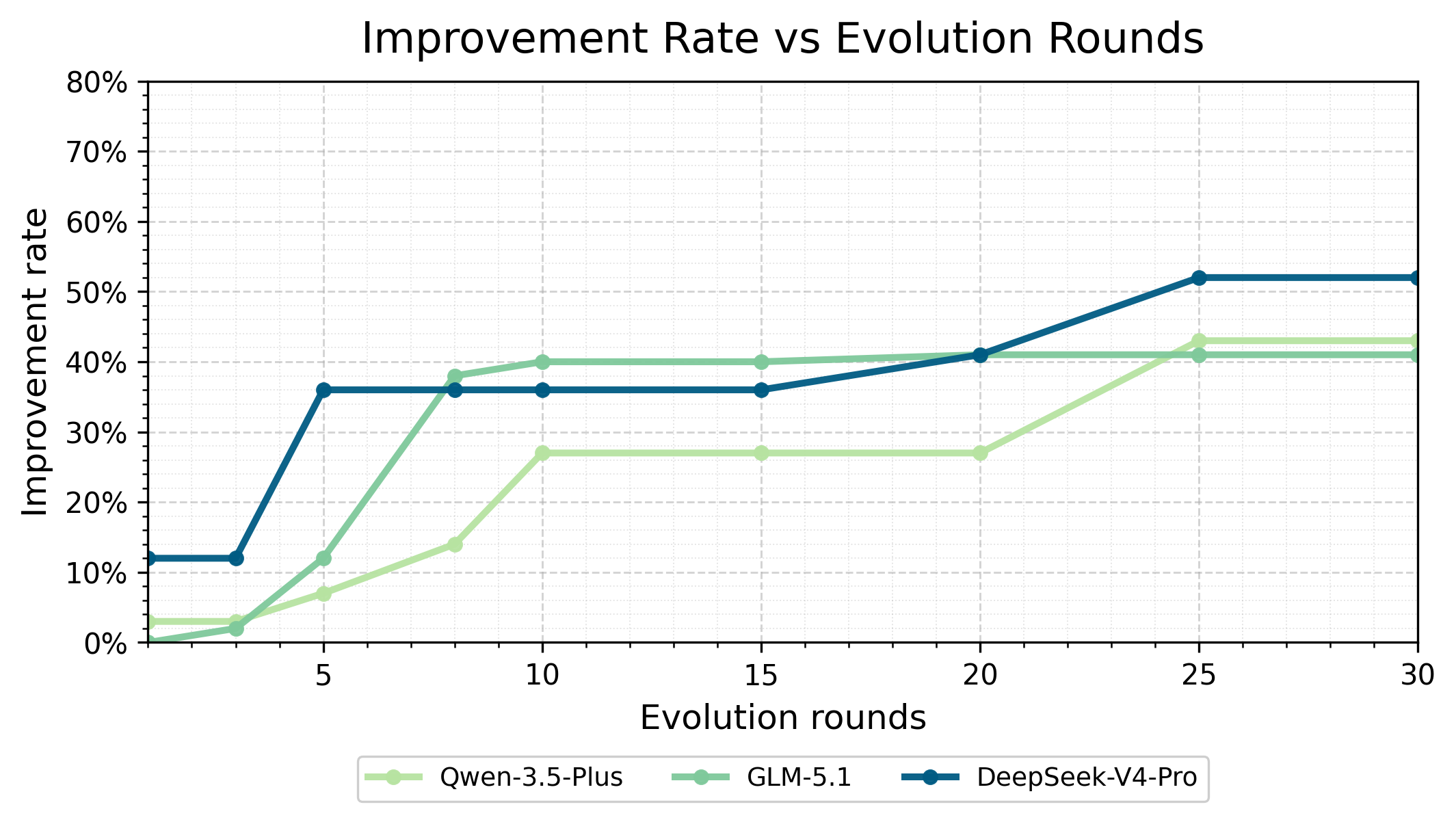}
    \caption{Best-so-far improvement rates over evolution rounds using different LLM backbones.}
    \label{fig:llm}
\end{figure}

\subsection{Comparison with Search and LLM-Agent Baselines}

\begin{table}[t]
\centering
\small
\setlength{\tabcolsep}{1mm}
\begin{tabular}{llcccc}
\toprule
Dim. & PDE
& Bayesian
& Lang-PINN
& PINNacle
& ADSL-PDE
\tabularnewline

\midrule

\multirow{2}{*}{1D}
& Burgers
& 8.70E-02
& 6.48E-05
& 7.90E-05
& \textbf{5.52E-05}
\tabularnewline

& KS
& 1.10E+00
& \textbf{1.62E-03}
& 1.04E+00
& 1.66E-03
\tabularnewline

\midrule

\multirow{6}{*}{2D}
& Burgers-C
& 2.42E-01
& \textbf{2.88E-03}
& 1.09E-01
& 1.28E-02
\tabularnewline

& Heat-CG
& 1.17E-01
& 1.35E-03
& 8.53E-04
& \textbf{3.21E-04}
\tabularnewline

& NS-C
& 5.12E-03
& 4.05E-05
& 2.33E-05
& \textbf{1.21E-05}
\tabularnewline

& Heat-MS
& 7.48E-03
& 2.27E-05
& 5.27E-05
& \textbf{2.25E-05}
\tabularnewline

& Heat-VC
& 3.93E-02
& 1.62E-03
& 1.76E-03
& \textbf{5.66E-04}
\tabularnewline

& Poisson-MA
& 5.82E+00
& 2.25E-03
& 1.83E+00
& \textbf{1.24E-04}
\tabularnewline

\midrule

3D
& Poisson-CG
& 2.55E-02
& 1.35E-03
& 9.51E-04
& \textbf{1.55E-04}
\tabularnewline

\midrule

\multirow{2}{*}{ND}
& Poisson-ND
& 4.72E-05
& 8.42E-06
& \textbf{2.09E-06}
& 1.55E-05
\tabularnewline

& Heat-ND
& 1.18E-04
& 4.72E-04
& 8.52E+00
& \textbf{9.13E-05}
\tabularnewline

\midrule

\multicolumn{2}{l}{Geo. Mean}
& 2.58E-02
& 3.41E-04
& 3.30E-03
& \textbf{1.58E-04}
\tabularnewline

\bottomrule
\end{tabular}
\caption{Comparison with search-based and prompt-driven LLM-agent baselines.
The reported metric is MSE or relative $L_2$ error; lower is better.
Geo. Mean summarizes performance across all evaluated tasks.}
\label{tab:search_agent_baselines}
\end{table}

Table~\ref{tab:search_agent_baselines} compares ADSL-PDE with search-based and prompt-driven LLM-agent baselines. ADSL-PDE achieves the best result on 8 out of 11 tasks and obtains the lowest geometric mean error across all methods. Compared with RandomAgent and BayesianAgent, ADSL-PDE consistently achieves much lower errors, showing that the improvement is not merely due to exploring more candidates. Instead, the structured DSL provides a more informative and better constrained search space.

Compared with prompt-driven LLM-agent baselines, ADSL-PDE also achieves stronger overall performance. For instance, on Poisson-MA, ADSL-PDE reduces the error to $1.24\times10^{-4}$, while PINNsAgent and PINNacle obtain $3.16$ and $1.83$, respectively. ADSL-PDE does not dominate every individual task: Lang-PINN performs best on KS and Burgers-C, while PINNacle performs best on Poisson-ND. Nevertheless, the overall geometric mean shows that ADSL-PDE provides the strongest average performance across heterogeneous PDE settings. This supports the central claim that a structured agent-facing language improves the reliability and effectiveness of solver auto-design.

\subsection{Sensitivity to LLM Backbone}

We evaluate ADSL-PDE with Qwen-3.5-Plus, GLM-5.1, and
DeepSeek-V4-Pro under the same search budget and backend protocol.
As shown in Figure~\ref{fig:llm}, all three models produce
consistent improvements as evolution proceeds. DeepSeek-V4-Pro achieves
the highest final improvement rate of approximately 52\%, while
Qwen-3.5-Plus and GLM-5.1 reach approximately 42\% and 41\%,
respectively. Although the final values differ, all backbones discover
substantially improved solvers and exhibit similar overall convergence
patterns.

These results suggest that the LLM backbone has a moderate, rather than
dominant, influence on ADSL-PDE. By restricting generation to structured
and valid solver-level decisions, the language system reduces the
dependence of the evolutionary process on a particular LLM. Therefore,
when a modest difference in final performance is acceptable, the LLM
backbone can be selected primarily according to inference cost and
availability. A stronger model may still be preferred when maximizing
the final solver performance is the primary objective.

\subsection{Ablation of Key Designs}

We evaluate the key components of ADSL-PDE by comparing direct Python
generation, structured specification, validated IR, backend support, and
the complete system. Table~\ref{tab:key_design_ablation} reports the
average performance improvement per 100K tokens and the candidate pass
rate under the same search budget.

\begin{table}[t]
\centering
\small
\setlength{\tabcolsep}{4.5mm}
\begin{tabular}{lcc}
\toprule
\textbf{Design}
& \textbf{\shortstack{Improvement per\\100K tokens}}
& \textbf{Pass rate} \\
\midrule
Python only
& 12.10\%
& 14/30 \\

Structured schema
& 14.74\%
& 13/30 \\

Validated IR
& 7.71\%
& \textbf{23/30} \\

Backend
& 21.62\%
& 19/30 \\

ADSL-PDE
& \textbf{30.18\%}
& \textbf{23/30} \\
\bottomrule
\end{tabular}
\caption{Ablation study of the key designs in ADSL-PDE. Improvement
per 100K tokens measures the average performance gain achieved for
every 100K generated tokens. Pass rate denotes the number of valid
candidates among 30 generated candidates. Higher values are better,
and the best results are shown in bold.}
\label{tab:key_design_ablation}
\end{table}

\begin{table}[t]
\centering
\small
\setlength{\tabcolsep}{1.0mm}
\begin{tabular}{@{}lcccc@{}}
\toprule
\textbf{Metric}
& \textbf{\shortstack{Direct\\Python}}
& \textbf{\shortstack{Plain\\Config}}
& \textbf{\shortstack{Low-level\\DSL}}
& \textbf{\shortstack{Full\\ADSL}} \\
\midrule
Burgers
& 1.83E-02
& 8.74E-03
& 2.36E-03
& \best{5.52E-05} \\

Burgers-C
& 1.11E-01
& 1.08E-01
& 6.12E-02
& \best{1.28E-02} \\

NS-C
& 1.30E-03
& 7.28E-05
& 3.01E-05
& \best{1.21E-05} \\

Poisson-ND
& 1.26E-04
& 1.02E-04
& 6.23E-05
& \best{1.55E-05} \\

Geo. Mean
& 4.27E-03
& 1.63E-03
& 7.21E-04
& \best{1.07E-04} \\

\midrule
VCR$\uparrow$
& 0.34
& 0.62
& 0.78
& \best{0.91} \\

SE$\uparrow$
& 0.18
& 0.31
& 0.46
& \best{0.83} \\

Token-SE$\uparrow$
& 0.06
& 0.15
& 0.31
& \best{0.72} \\
\bottomrule
\end{tabular}
\caption{Search-space ablation evaluating the semantic compression
effect of ADSL-PDE. Task performance is measured by MSE, for which
lower values are better. VCR, SE, and Token-SE measure search
reliability and efficiency, for which higher values are better.}
\label{tab:search_space_ablation}
\end{table}

\begin{table}[t]
\centering
\small
\setlength{\tabcolsep}{3.0mm}
\begin{tabular}{@{}lcc@{}}
\toprule
\textbf{Task}
& \textbf{Base}
& \textbf{+ ADSL-PDE} \\
\midrule

\multicolumn{3}{@{}l}{\textit{OpenEvolve}} \\
Burgers
& $1.14\times10^{-2}$
& \improved{$5.52\times10^{-5}$} \\

Burgers-C
& $4.36\times10^{-1}$
& \best{\improved{$1.28\times10^{-2}$}} \\

NS-C
& $7.91\times10^{-4}$
& \improved{$1.21\times10^{-5}$} \\

Poisson-MA
& $3.74\times10^{-2}$
& \best{\improved{$1.24\times10^{-4}$}} \\

Poisson-ND
& $3.22\times10^{-3}$
& \best{\improved{$1.55\times10^{-5}$}} \\

\midrule
\multicolumn{3}{@{}l}{\textit{PINNsAgent}} \\
Burgers
& $6.51\times10^{-5}$
& \best{\improved{$3.12\times10^{-5}$}} \\

Burgers-C
& $2.04\times10^{-1}$
& $2.09\times10^{-1}$ \\

NS-C
& $8.50\times10^{-6}$
& \best{\improved{$1.70\times10^{-6}$}} \\

Poisson-MA
& $3.16\times10^{0}$
& \improved{$8.13\times10^{-2}$} \\

Poisson-ND
& $4.77\times10^{-4}$
& \improved{$2.96\times10^{-5}$} \\

\midrule
\multicolumn{3}{@{}l}{\textit{OpenAlpha-Evolve}} \\
Burgers
& $1.22\times10^{-3}$
& \improved{$3.18\times10^{-4}$} \\

Burgers-C
& $2.47\times10^{-1}$
& $2.89\times10^{-1}$ \\

\bottomrule
\end{tabular}
\caption{Performance of different evolution backends with and without
ADSL-PDE. Results are reported in MSE, and lower values indicate better
performance. Base denotes the original evolution backend. The best
result on each task is shown in bold, and $\uparrow$ indicates that
ADSL-PDE improves upon the corresponding base backend. ``--'' denotes
unavailable results.}
\label{tab:evolution_backend_comparison}
\end{table}

Direct Python generation achieves an average improvement of 12.10\% per
100K tokens, with only 14 of 30 candidates passing the validity checks.
Structured specification slightly improves token efficiency but does
not increase the pass rate, indicating that a structured format alone
cannot ensure executable solver designs. Introducing validated IR raises
the pass rate to 23/30 by detecting invalid candidates before execution,
although validation alone does not directly improve solver quality.
Backend support increases the average improvement to 21.62\%, showing
the value of mapping solver specifications to reliable and reusable
implementations.

The complete ADSL-PDE achieves the highest token-normalized improvement
of 30.18\% while retaining the joint-best pass rate of 23/30. These
results indicate that structured representation, IR validation, and
backend compilation provide complementary benefits. Their integration
improves both the reliability of generated candidates and the efficiency
of neural PDE solver evolution.

\subsection{Representation Ablation}

We evaluate whether the language representation improves solver evolution
by removing unproductive regions from the agent's search space.
Table~\ref{tab:search_space_ablation} compares direct Python generation,
plain configuration search, low-level DSL search, and full ADSL-PDE under
the same PDE tasks, search budget, and backend protocol. These
representations progressively reduce implementation-level freedom while
preserving solver-level design choices.

Full ADSL-PDE achieves the lowest geometric mean error and the highest
VCR, SE, and Token-SE. Compared with direct Python generation, it reduces
the geometric mean error from $4.27\times10^{-3}$ to
$1.07\times10^{-4}$, increases VCR from $0.34$ to $0.91$, and improves
SE and Token-SE from $0.18$ and $0.06$ to $0.83$ and $0.72$,
respectively. These results show that ADSL-PDE concentrates exploration
on valid and performance-relevant solver decisions, thereby improving
search efficiency.

Figure~\ref{fig:combined_scores} further shows that DSL-based evolution
improves faster and fluctuates less than direct code generation. Overall,
the representation ablation confirms that reducing irrelevant
implementation choices improves both the efficiency and reliability of
neural PDE solver evolution.

\subsection{Generality Across Evolution Backends}

ADSL-PDE is designed as a representation layer that can be combined with
different evolutionary search methods. We therefore compare OpenEvolve,
PINNsAgent, and OpenAlpha-Evolve with and without ADSL-PDE on the same
PDE tasks. The results are reported in
Table~\ref{tab:evolution_backend_comparison}.

ADSL-PDE improves 10 of the 12 available task--backend comparisons.
The improvement is most consistent with OpenEvolve, for which ADSL-PDE
reduces the error on all five tasks. When combined with PINNsAgent,
ADSL-PDE improves four of the five tasks, with only a slight degradation
on Burgers-C. For OpenAlpha-Evolve, it improves Burgers but does not
improve Burgers-C.

These results indicate that the effectiveness of ADSL-PDE is not tied to
a specific evolution algorithm. By providing a structured and constrained
solver representation, ADSL-PDE can generally improve the search
performance of different evolutionary backends. Nevertheless, the gains
are task-dependent and are not guaranteed for every backend--task
combination.

\section{Discussion}

Our results show that ADSL-PDE is an effective representation layer for neural PDE solver auto-design. Its main benefit come from reshaping the search space. By replacing unrestricted Python generation with structured solver-level decisions,
ADSL-PDE removes many invalid or implementation-irrelevant candidates and concentrates agent exploration on choices that directly affect solver performance. This leads to higher candidate validity, better token efficiency,
and more stable evolution across different PDE tasks, LLM backbones, and evolutionary frameworks.

The study also suggests several principles for designing effective agent-oriented languages. First, the language should expose domain-relevant decisions, such as solver family, architecture, physical constraints, sampling, and training, while hiding low-level implementation details. Second, it should provide controlled expressiveness: overly rigid representations restrict innovation, whereas overly general code spaces introduce unnecessary search dimensions. Third, static validation and deterministic compilation are
essential for rejecting invalid candidates before expensive training and for ensuring reproducibility. 

These findings support the central claim of this work: improving neural PDE solver auto-design requires not only stronger agents, but also a language system that defines a compact, valid, and semantically effective search space. ADSL-PDE provides one realization of this principle and offers a general direction for language-guided scientific solver evolution.
Future work will systematically evaluate how language systems affect LLM-agent performance across diverse tasks and theoretically derive bounds on their improvements in task completion.

\bibliography{aaai2027}
\end{document}